# PART OF SPEECH TAGGING OF MARATHI TEXT USING TRIGRAM METHOD

Jyoti Singh[1], Nisheeth Joshi[2], Iti Mathur[3]

[1,2,3]Apaji Institute, Banasthali University, Rajasthan, India,
[1]*jyoti.singh132@gmail.com,*
[2]*nisheeth.joshi@rediffmail.com,*
[3] *mathur_iti@rediffmail.com*

*ABSTRACT*

*In this paper we present a Marathipart of speech tagger. It is morphologically rich language. it is spoken by the native people of Maharashtra. The general approach used for development of tagger is statistical using Trigram Method. The main concept of Trigram is to explore the most likely POS for a token based on given information of previous two tags by calculating probabilities to determine whichthe best sequence of tag is. In this paper we show the development of the tagger. Moreover we have also shown the evaluation done.*

*KEYWORDS:*

*Part of Speech Tagging, Stochastic Tagging, N-gram Modeling,Marathi*

## 1. INTRODUCTION

Natural Language is the medium for communication which is incorporated by every human being. One of the most important activities in processing natural languages is Part of Speech tagging. In POS Tagging we assign a Part of Speech tag to each word in a sentence and literature. POS tagging is one of the simplest, most constant and statistical model for many NLP application. POS Tagging is an initial stage of linguistics, text analysis like information retrieval, machine translator, text to speech synthesis, information extraction etc. Part-of-speech tagging is a process of assigning the words in a text as corresponding to a particular part of speech.

A fundamental version of POS tagging is the identification of words as nouns, verbs, adjectives etc. POS Tagging can be regarded as a simplified form of morphological analysis where it only deals with assigning an appropriate POS tag to the word, while morphological analysis deals with finding the internal structure of the word. Indian languages are morphologically rich and they have more than one morpheme of a word due to this tagging of Indian languages are difficult.A Part-Of-Speech Tagger (POS Tagger) is a piece of software that reads text in some language and assigns parts of speech to each word. A part of speech includes nouns, verbs, adverbs, adjectives, pronouns, conjunction and their sub-categories. Various approaches have been proposed to International Journal of Advanced Information Technology (IJAIT) Vol. 3, No.2, April2013implement POS taggers. Broadly they can be classified as rule based Statistical and Hybrid approaches.The rule based POS tagging models apply a set of hand written rules and use contextual information to assign POS tags to words.

The necessity of a linguistic background and manually constructing the rules are the main drawbacks of the rule based systems.A stochastic approach includes frequency and probability or statistics. The simplest stochastic approach finds out the most frequently used tag for a specific word in the annotated training data and uses this information to tag that word in the unannotated text. The problem with this approach is that it can come up with sequences of tags for sentences that are not acceptable according to the grammar rules of a language.The Hybrid approaches use a pre-defined set of handcrafted rules as well as automatically induced rules that are generated during training.

## 2. PROBLEMS OF PART OF SPEECH TAGGING

Ambiguous words are the main problem in part of speech tagging. There may be many words which can have more than one tag. Sometimes it happens that a word has same POS but have different meaning in different context. To solve this problem we consider the context instead of taking single word. For example-

/**NNP** /PSP,/SYM /PRP /VM /**RB** /VAUX./SYM

The same word 'अंकुश' is given a different label in a same sentence. In the first case it is termed as a proper noun. In the second case it is termed as a common noun as it is referring to control. Since first word अंकुश occur in a sentence as subject and after that there is a postposition that's why it is labelled as NNP and in second time अंकुश comes between main verb and auxiliary verb so it is assigned as adverb. POS Tagging tries to correctly identify a POS of a word by looking at the context (surrounding words) in a sentence.

## 3. PREVIOUS WORK ON INDIAN LANGUAGE POS TAGGING

There are several approaches that have been used for POStagging and several research workhave been carried out in this area. Singh et. al. [2] proposed a Manipuri POS Tagger using CRF and SVM. In this paper, they described a tagger for using Conditional Random Field (CRF) and Support Vector Machine (SVM). There Evaluation results demonstrated the accuracies of 72.04%, and 74.38% in the CRF, and SVM, respectively. Hasanet. al. [3] Proposed Comparison of Unigram, Bigram, HMM and Brill's POS Tagging Approaches for some South Asian Languages. They used stochastic methodology for some of south Asian languages like Bangla, Hindi and Telugu and compared corresponding result for all languages.They found that brill transformation based taggers results are better. Ekbal and Sivaji [4] proposed Web-based Bengali News Corpus for Lexicon Development and POS Tagging the POS taggers using Hidden Markov Model (HMM) and Support Vector Machine (SVM). The POS taggers are developed for Bengali shows the accuracies as 85.56%, and 91.23% for HMM, and SVM, respectively.

Dhanalakshmi V,et. al. [5] presentedTamil POS Tagging using Linear Programming. In this paper they proposed the Part Of Speech tagger for Tamil using Machine learning techniques. They

36

discovered that SVM based machine learning tool affords the most encouraging result for Tamil POS tagger (95.64%). Kumaret. al. [6] presentedBuilding Feature Rich POS Tagger for Morphologically Rich Languages: Experiences in Hindi. A statistical part-of-speech tagger for a morphologically rich language: Hindi.This Tagger employs the maximum entropy Markov model with a rich set of features capturing the lexical and morphological characteristics of the language. The system achieved the best accuracy of 94.89% and an average accuracy of 94.38%.Singhet. al. [7], in 2008 proposed Part-of-Speech Tagging for Grammar Checking of Punjabi. In this paper, they have discussed the issues concerning the development of a POS tagset and a POS tagger for the use as a part of the project on developing an automated grammar checking system for Punjabi Language.In 2009, Manju K.et. al. [8] proposed Development of a POS Tagger for Malayalam which was a Hidden Markov Model (HMM) based part of speech tagger for Malayalam language. The performance of the developed POS Tagger is about 90% and almost 80% of the sequences that have been generated automatically for the test case were found correct. Joshiet. al. [9], proposed Part of Speech Tagging for Hindi. They have used IL POS tag set for the development of the tagger. They disambiguated correct word-tag combinations using the contextual information available in the text.They have achieved the accuracy of 92%.

Patel et. al. [10] proposed Part-Of-Speech Tagging for Gujarati Using Conditional Random Fields. In this paper they described a machine learning algorithm for Gujarati Part of Speech Tagging.This paper shows a machine learning algorithm for Gujarati Part of Speech Tagging. The machine learning part is performed using a CRF model.The algorithms have achieved an accuracy of 92% for Gujarati texts where the training corpus is of 10,000 words and the test corpus is of 5,000 words. Reddy and Sharoff[11] proposed Cross Language POS Taggers (and other Tools) for Indian Languages: An Experiment with Kannada using Telugu Resources they have usedTnT (Brants, 2000), a popular implementation of the second-order Markov model for POS tagging. Kumar et. al[12] presented Part of Speech Tagger for Morphologically rich Indian Languages: A survey. In this paper they have reported about differentPOS taggers based on different languages and methods. In this paper they have shown the accuracy of corresponding tagger.

## 4. POS TAGSET

Depending on some general principle of tagset design strategy, a number of POS tagsets have been developed by different organizations based. For POS annotation of texts in Marathi, we have used tagset developed by IIIT Hyderabad (Bharti, et. al., 2006) [1]. Table shows brief description of IL POS Tag set.

| S.No. | Tag | Description (Tag Used for) | Example |
|---|---|---|---|
| 1. | NN | Common Nouns | , , , |
| 2. | NST | Noun Denoting Spatial and Temporal Expressions | , , |
| 3. | NNP | Proper Nouns (name of person) | , , |
| 4. | PRP | Pronoun | , , |
| 5. | DEM | Demonstrative | , , , , |
| 6. | VM | Verb Main (Finite or Non-Finite) | , , |
| 7. | VAUX | Verb Auxiliary (Any verb, present besides main verb shall be marked as auxiliary verb) | , , , , |
| 8. | JJ | Adjective (Modifier of Noun) | , , |
| 9. | RB | Adverb (Modifier of Verb) | , , , |
| 10. | PSP | Postposition | , , |
| 11. | RP | Particles | , , |
| 12. | QF | Quantifiers | , , |
| 13. | QC | Cardinals | , , |
| 14. | CC | Conjuncts (Coordinating and Subordinating) | , , , , |
| 15. | WQ | Question Words | |
| 16. | QO | Ordinals | |
| 17. | INTF | Intensifier | |
| 18. | INJ | Interjection | , , , |
| 19. | NEG | Negative | , |
| 20. | SYM | Symbol | ? , ; : ! |
| 21. | XC | Compounds | - , - |
| 22. | RDP | Reduplications | - |
| 24. | UNK | Foreign Words | English, , |

Table 1: POS tagset for Marathi

## 5. OUR APPROACH

In this paper we are describing Trigram Model for Marathi POS tagger. Our main aim is to perform POS Tagging to determine the most likely tag for a word, given the previous two tags. For trigrams, the probability of a sequence is just the product of conditional probabilities of its trigrams. So if $t_1, t_2 \ldots t_n$ are tag sequence and $w_1, w_2 \ldots w_n$ are corresponding word sequence then the following equation explains this fact-

$$P(t_i|w_i) = P(w_i|t_i).P(t_i|t_{i-2,}t_{i-1}) \tag{1}$$

Where $t_i$ denotes the tag sequence and $w_i$ denotes the word sequences
$P(w_i|t_i)$ is the probability of current word given current tag.

Here, $P(t_i|t_{i-2}t_{i-1})$ is the probability of a current tag given the previous two tags.

This provides the transition between the tags and helps capture the context of the sentence. These probabilities are computed by following equation.

$$P(t_i/t_{i-2}, t_{i-1}) = f(t_{i-2}, t_{i-1}, t_i)/f(t_{i-2}, t_{i-1}) \tag{2}$$

Each tag transition probability is computed by calculating the frequency count of two tags which come together in the corpus divided by the frequency count of the previous two tagscomingin the corpus.

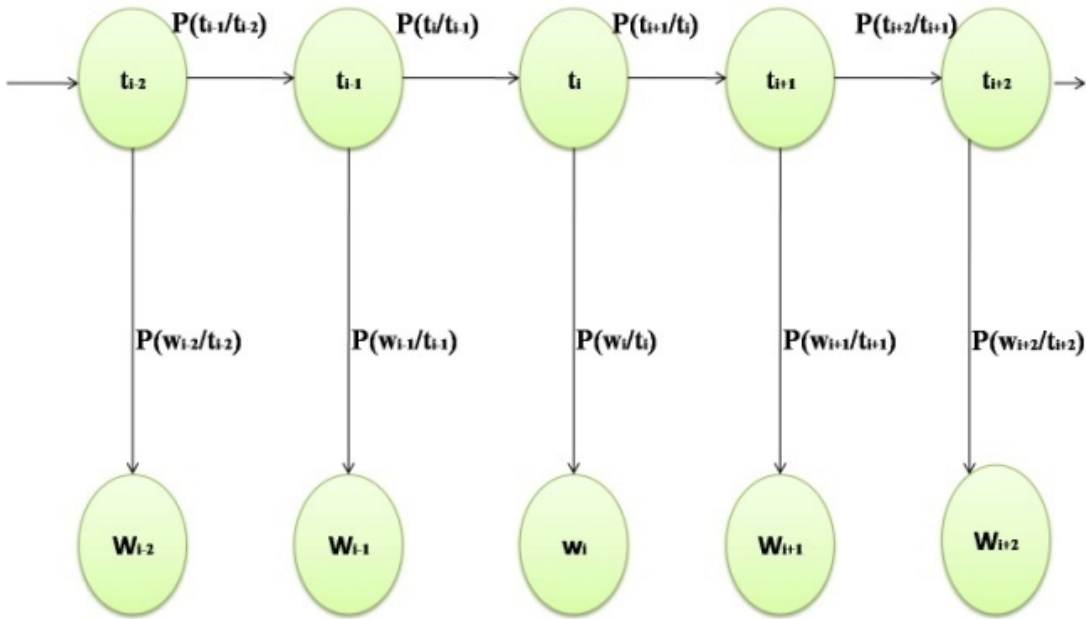

Figure 1. Working of trigram model

International Journal of Advanced Information Technology (IJAIT) Vol. 3, No.2, April2013

## 6. EVALUATION

For testing the performance of our system, we developed a test corpus of 2000 sentences (48,635 words). This provided us with accuracy of 91.63%. The accuracy was calculated using the formula:

$$\text{Accuracy}(\%) = \frac{\text{No. of correctly Tagged tokens}}{\text{Total No. of POS tags in the text}} * 100 \qquad (3)$$

Test scores of our system are as follows:
No. of Correct POS tags assigned bythe system = 44563
No. of POS taginthe text = 48635
Thus the accuracy of the system is 91.63%.

## 7. CONCLUSION

The topic of POS tagging discussed in this paper showsTrigram based POS tagger for Marathi. The POS tagger described here is very simple and efficient for automatic tagging, but the morphological complexity of the Marathi makes it a little hard. The performance of the current system is good and the results achieved by this method are excellent. In future we wish to improve the accuracy of our system by adding more tagged sentences in our training corpus.

## AUTHORS

Jyoti Singh is pursuing her M.Tech in Computer Science from Banasthali University, Rajasthan and is working as a Research Assistant in English-Indian Languages Machine Translation System Project sponsored by TDIL Programme, DeitY. She has her interest in Machine Translation specifically in English-Marathi Language Pair. She has developed various tools on Marathi Language Processing. Her current research interest includes Natural Language Processing and Machine Translation.

Nisheeth Joshi is an Assistnat Professor at Banasthali University. He has been primarily working in design and development of evaluation Matrices in Indian languages. Besides this he is also actively involved in the development of MT engines for English to Indian Languages. He is one of the experts empanelled with TDIL programme, Department of Electronics and Information Technology (DeitY), Govt. of India, a premier organization which foresees Language Technology Funding and Research in India. He has several publications in various journals and conferences and also serves on the Programme Committees and Editorial Boards of several conferences and journals.

Iti Mathur is an Assistant Professor at Banasthali University. Her primary area of research is Computational Semantics and Ontological Engineering. Besides this she is also involved in the development of MT engines for English to Indian Languages. She is one of the experts empanelled with TDIL Programme, Department of Electronics and Information Technology (DeitY), Govt. of India, a premier organization which foresees Language Technology Funding and Research in India. She has several publications in various journals and conferences and also serves on the Programme Committees and Editorial Boards of several conferences and journals.